\documentclass[conference]{IEEEtran}
\IEEEoverridecommandlockouts
\usepackage{cite}
\usepackage{amsmath,amssymb,amsfonts}
\usepackage{algorithmic}
\usepackage{graphicx}
\usepackage{textcomp}
\usepackage{xcolor}

\usepackage{multirow}
\usepackage{multicol}
\usepackage[linesnumbered, boxed, ruled]{algorithm2e}
\usepackage{wrapfig}
\usepackage{makecell}
\usepackage{booktabs} 
\usepackage{url}
\definecolor{lightred}{RGB}{251,49,153}

\def\BibTeX{{\rm B\kern-.05em{\sc i\kern-.025em b}\kern-.08em
    T\kern-.1667em\lower.7ex\hbox{E}\kern-.125emX}}
\begin{document}

\title{Map-Free Visual Relocalization Enhanced by Instance Knowledge and Depth Knowledge\\
% {\footnotesize \textsuperscript{*}Note: Sub-titles are not captured for https://ieeexplore.ieee.org  and
% should not be used}
% \thanks{Identify applicable funding agency here. If none, delete this.}
}

\author{\IEEEauthorblockN{Mingyu Xiao\textsuperscript{1,2*}, Runze Chen\textsuperscript{1,2*}, Haiyong Luo\textsuperscript{2$\dagger$}, Fang Zhao\textsuperscript{1$\dagger$}, Fan Wu\textsuperscript{1,2}, Hao Xiong\textsuperscript{1,2}, Juan Wang\textsuperscript{3}, Xuepeng Ma\textsuperscript{3}}
\IEEEauthorblockA{
\textsuperscript{1}\textit{Beijing University of Posts and Telecommunications, Bejing, China}\\
\textsuperscript{2}\textit{Institute of Computing Technology, Chinese Academy of Sciences, Bejing, China}\\
\textsuperscript{3}\textit{Shouguang Cheng Zhi Feng Xing Technology Co., Ltd., ShanDong, China}\\
\{shawnmy,chenrz925,zfsse,wufan98326,xmr2015211989\}@bupt.edu.cn, yhluo@ict.ac.cn, sllhjt@139.com,139225785@qq.com}

\thanks{* These authors contributed equally to this work.}
\thanks{$\dagger$ Corresponding authors: Haiyong Luo and Fang Zhao.}

}

\maketitle

\begin{abstract}

Map-free visual relocalization computes camera pose using only a query image and a reference image.
Therefore, it is hindered by challenges in feature-point matching and the absence of scale information in monocular images.
These issues may cause significant rotational and metric errors, leading to localization failures.
To address these challenges, we propose a map-free visual relocalization method enhanced with instance knowledge and depth knowledge. 
By utilizing instance-based matching, our approach improves the robustness of feature-point matching by focusing on relevant regions across scenes. 
Additionally, our depth estimation techniques provide reliable depth knowledge from a single image, improving scale recovery and reducing translation errors.
Our method surpasses the previous state-of-the-art by 1.071m and 13.593° on the translation error and rotation error, respectively. Furthermore, we are also one of the winners in the Map-free Workshop \& Challenge (ECCV2024), underscoring its superiority compared to concurrent approaches.

% The proposed method won third place in the Map-Free Visual Relocalization Workshop \& Challenge (ECCV 2024).

\end{abstract}

\begin{IEEEkeywords}
Map-Free Visual Relocalization, Instance Knowledge, Depth Knowledge
\end{IEEEkeywords}

\section{Introduction}
Visual relocalization estimates camera position and orientation from a query image, which offers numerous applications such as augmented reality and robotic navigation. Early map-based approaches~\cite{svarm2016city,li2012worldwide,brachmann2021visual,panek2023visual,kim2023ep2p} rely on image retrieval to estimate poses. 
Specifically, some methods~\cite{abs-2103-11468,LaskarMKK17,kendall2016modelling,brahmbhatt2018geometry} designs end-to-end models. Other methods utilize modules that match features~\cite{Sun_2021_CVPR,sarlin2020superglue,lowe2004distinctive}, depth estimation~\cite{wu2023depth,miao2018active,lo2020depth} and estimate poses successively. 
These methods require large image datasets and detailed 3D maps, limiting scalability in memory-constrained environments. Map-free visual relocalization~\cite{arnold2022map}, by contrast, computes camera pose using only a query image and a reference image, eliminating the need for pre-built maps.
However, it faces challenges due to limitations in feature-point matching and the lack of scale information in monocular images.

\begin{figure}[h]
% \vspace{-4.5mm}
\begin{center}
\includegraphics[width=\linewidth]{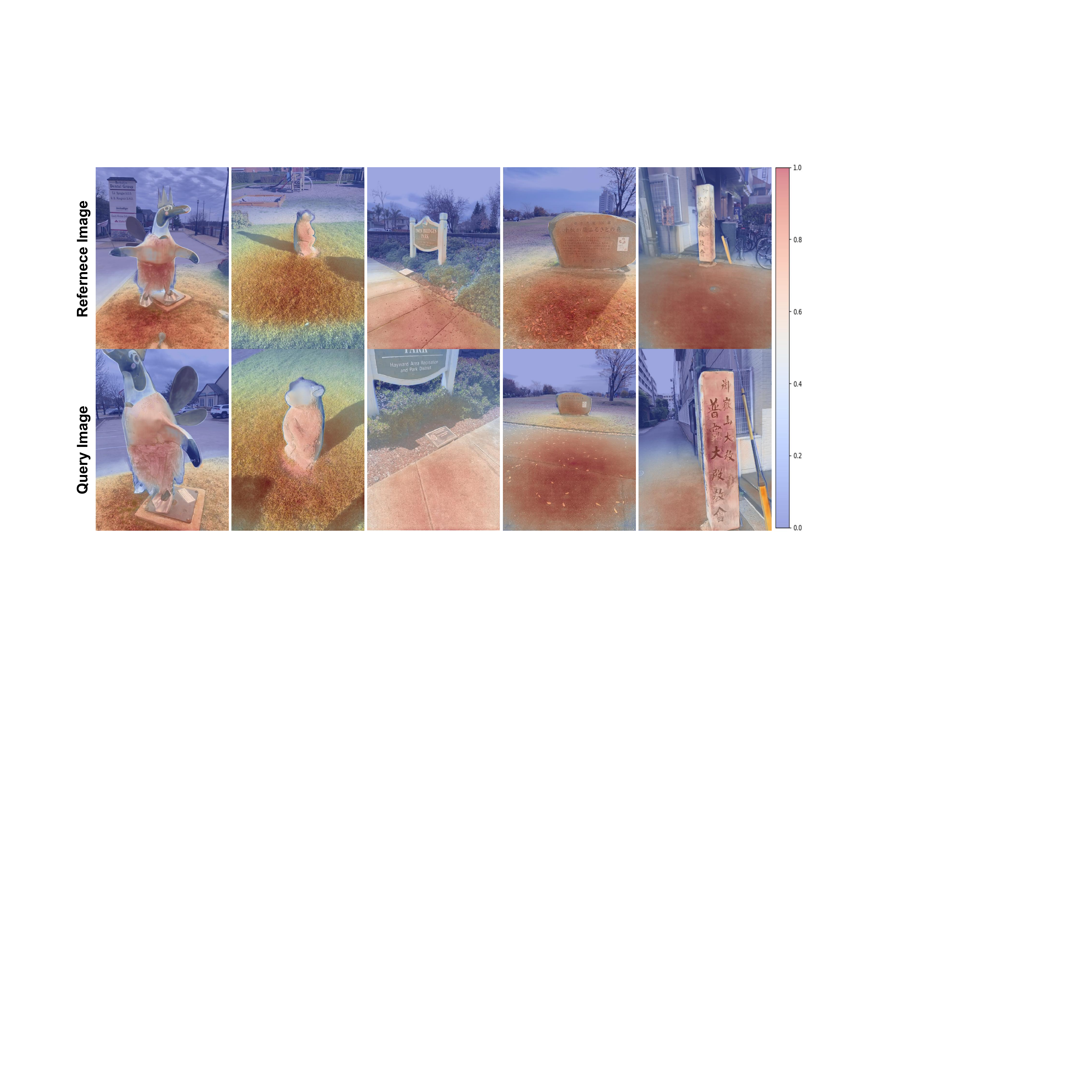}
\end{center}
% \vspace{-2mm}
\caption{
The matched points foreground objects exhibit slightly higher confidence, but this preference is not significant.
}
\label{fig: object}
% \vspace{-4mm}
\end{figure}

To improve the accuracy of map-free relocalization, we propose a novel framework that not only enhances matching precision but also effectively addresses the challenges of metric depth estimation. Our approach significantly advances visual relocalization by systematically reducing both rotation and translation errors, which are critical for achieving reliable localization without the need for pre-existing maps. 
To reduce the rotation errors, we delve into a powerful Muli-view stereo reconstruction model DUSt3R~\cite{wang2024dust3r}, which can obtain pairwise pointmaps based on the extracted 3D features from two images. 
Despite DUSt3R’s good matching ability, its performance on map-free visual relocalization remains suboptimal. To investigate this, we visualize the confidence of the pointmaps between the reference and query frames in Figure~\ref{fig: object}. Points closer to red indicate higher confidence, while those nearer to blue show lower confidence. As illustrated in Figure~\ref{fig: object}, some foreground objects exhibit slightly higher confidence, but this preference is not significant. However, in map-free visual relocalization tasks, accurate matching of foreground objects is crucial for successful relocalization, because foreground objects are more robust features in the global scene, making them more conducive to precise relocalization. The lack of significant confidence in the foreground object areas explains why DUSt3R performs well in multi-view stereo reconstruction but struggles with map-free visual relocalization.
Thereby, to fully utilize the matching ability of DUSt3R, we introduce a feature point matching method enhanced by instance knowledge, which restricts the matching scope to within specific instances. By guiding the model to focus on specific instance matches, we can effectively reduce incorrect matches between different objects.

\begin{figure*}[t]
\begin{center}
\includegraphics[width=\linewidth]{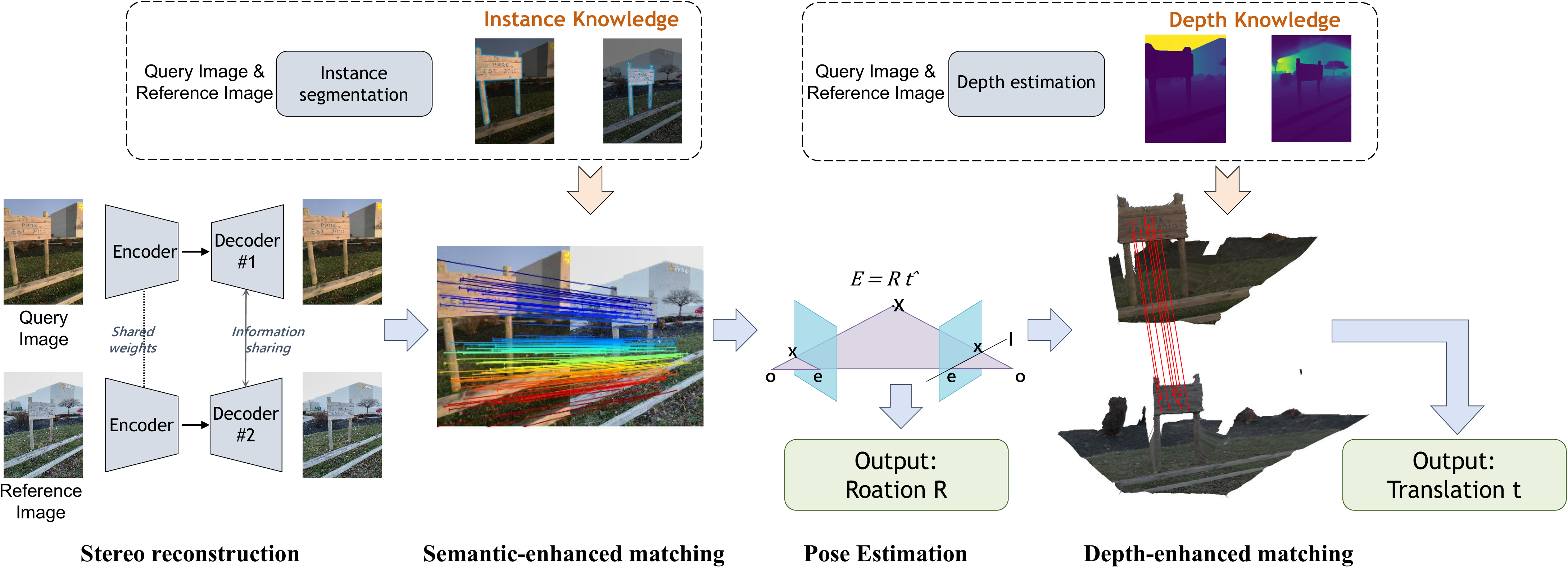}
\end{center}
% \vspace{-2mm}
\caption{Overview. Our model first takes two input images as input and obtains aligned point maps. Then, utilizing instance segmentation knowledge, feature points are matched both globally and within identified masks. The matched points are then input into an essential matrix solver, which computes the rotation matrix $R$ and a scale-free translation vector $\bar{t}$. Subsequently, depth knowledge is applied to project the feature points into 3D space, allowing for the recovery of a scaled translation vector $t$.} 
\label{fig:overview}
\end{figure*}

To reduce translation errors, our framework incorporates Metric3D~\cite{hu2024metric3d}, which employs standard camera transformations and joint depth-normal optimization to tackle the metric depth estimation problem from a single image. 
As our framework requires depth information to recover the scale of the translation vector,
accurate scale recovery can achieve better translation accuracy.
Our method achieves the best results on the map-free visual relocalization dataset compared to all methods mentioned in \cite{arnold2022map}, highlighting the effectiveness of our method.
Our contributions are as follows:
\begin{itemize}
    \item We proposed a hierarchical matching method that integrates instance-level and feature-level matching, effectively combining the advantages of global instance-level semantic matching and local feature point-based matching. This approach improves the accuracy of relocalization in map-less scenarios, with instance-level matching enhancing the precision of global alignment.
    \item  We conducted extensive experiments on complex scenarios within the map-free visual relocalization dataset, which includes spatiotemporal differences and significant parallax. These experiments demonstrate that our method exhibits superior generalization performance, surpassing existing algorithms.
\end{itemize}

\section{Method}

% \subsection{Overview}
The overall framework of our method is shown in Figure \ref{fig:overview}. 
We leverage instance knowledge to enhance the feature point matching and depth knowledge to facilitate scale recovery.

\subsection{Instance-enhanced Matching}
We utilize DUSt3R\cite{wang2024dust3r} to perform feature point matching by processing two input images \( I_1 \) and \( I_2 \), and outputs the corresponding point maps \( X_1 \) and \( X_2 \).
% For a pair of images $I_1, I_2$ and corresponding pointmaps $X_{1,1}, X_{2,1}$, we aim to find the matching point. 
Then, the nearest neighbor matching and reciprocal matching are utilized to find the matching point.
Firstly, we use the nearest neighbor search in the 3D point map space. For each pixel $(i,j)$ in image $I_1$, we first find the corresponding 3D point $X_{1,1}(i,j)$ in $X_{1,1}$. Then we perform a nearest neighbor search in $X_{2,1}$ to localize the 3D point $X_{2,1}(k,l)$ that is closest to $X_{1,1}(i,j)$, where $(k,l)$ are the coordinates of the matching point found in $X_{2,1}$. Then, we use reciprocal matching to improve the accuracy of the matching. From each pixel $(i,j)$ in $I_1$, we find the matching point $(k,l)$ in $I_2$. Simultaneously, from each pixel $(k,l)$ in $I_2$, we find the matching point $(i,j)$ in $I_1$. A match is considered valid only if the results are consistent in both directions, i.e., $(i,j) \rightarrow (k,l)$ and $(k,l) \rightarrow (i,j)$. We denote the global mapping as $Map^{g}=[((i_1,j_1),(k_1,l_1)), ((i_2,j_2),(k_2,l_2)), ...((i_n,j_n),(k_n,l_n))]$, where each tuple in the map represents the matching relation between points in $I_1$ and $I_2$, we define the $\mathcal{M}()$ as the feature point matching method:

\begin{equation}
    Map^{g}=\mathcal{M}(I_1,I_2,X_{1,1},X_{2,1}),
\end{equation}
where the $Map^{g}$ is the correspondences in global area.
To significantly enhance the precision of feature point matching, instance segmentation technology is integrated, enabling the accurate extraction of primary objects from images and focusing on matching within identical instance objects. 
Specifically, we use SegGPT~\cite{wang2023seggpt} to segment the same object in both the reference and query images. The masks for the $o$-th object are denoted as $Mask_{1}^{o}$ for the reference frame and $Mask_{2}^{o}$ for the query frame, respectively.
Then, our feature point matching algorithm conducts refined matching within these masks. 
Specifically, for the object $o$, we denote corresponding image as $I_{1}^{o}$ and $I_{2}^{o}$, where $I_{1}^{o}=I_1\bigodot Mask_{1}^{o}$, $I_{2}^{o}=I_2\bigodot Mask_{2}^{o}$.
Then, we perform point map reconstruction within the object and obtain corresponding point maps $X_{1,1}^{o}$ and $X_{2,1}^{o}$. Finally, we conduct matching within the mask object area:
\begin{equation}
    Map^{o}=\mathcal{M}(I_1^{o},I_2^{o},X_{1,1}^{o},X_{2,1}^{o}),
\label{map}
\end{equation}
where the $Map^{o}$ is the correspondences in the same object.
This modification aims for more precise matching within individual instances, ensuring that the feature-matching process is restricted to relevant areas defined by the instance masks. Finally, for areas of instance object, we use $Map^{o}$. For background areas, we use $Map^{g}$:
\begin{equation}
    Map=\left\{\begin{array}{l}
Map^{o} \quad\quad\quad if \ is \ instance \ object,\\ 
Map^{g} \quad\quad\quad else
\end{array}\right.
\label{eqmap}
\end{equation}

% Overall, the utilization of instance segmentation acts as a powerful contextual aid, enhancing the understanding of the context.

\subsection{Pose Estimation}
We selected the 5-point algorithm~\cite{nister2004efficient} combined with the essential matrix~\cite{hartley2003multiple} to compute the essential matrix $E$ from matched point pairs:
\begin{equation}
E=U\Sigma V^T,
\label{eqE1}
\end{equation}
where $U$ and $V$ are orthogonal matrices representing rotation or reflection, and $\Sigma$ is a diagonal matrix representing the original matrix's stretching factor.
The matrix $E$ is then decomposed into the rotation matrix $R$ and the scale-free translation vector $\bar{t}$:
% After this decomposition, additional geometric constraints are utilized to select the correct $R$ and $\bar{t}$:
\begin{equation}
E=\bar{t} \times R. 
\label{eqE2}
\end{equation}
% where $\bar{t}$ is the scale-free translation vector, and $R$ is the rotation matrix.

\subsection{Depth-enhanced Scale Recovery}
We use Metric3D \cite{hu2024metric3d} for scale recovery. The input image $I$ is initially transformed into a canonical camera space image $I_c$, which is then processed through an encoder-decoder network to predict a depth $d_c$. Finally, the predicted canonical depth $d_c$ is converted back to the actual depth $d$ in the original camera space by scaling $d_c$ using the ratio $\omega_d$, \textit{i.e.}, $d = \omega_d d_c$.

To integrate depth information into matching, we combine the depth estimation results $d_1, d_2$ with matching point pairs $Map$ from the instance-enhanced matching process. 
Assuming that the coordinates of a feature point pair in the reference image are $(i_1,j_1)$, and those of the corresponding feature point pair in the query image are $(k_1,l_1)$,  we can calculate their 3D coordinates assisted with depth values $d_1$ and $d_2$:
\begin{equation}
    {p}_1=d_1\cdot K^{-1}\begin{bmatrix}i_1\\j_1\\1\end{bmatrix}  ~~~{p}_2=d_2\cdot K^{-1}\begin{bmatrix}k_1\\l_1\\1\end{bmatrix},
\label{p}
\end{equation}
where $K$ represents the camera parameter matrix. ${p}_1$ and ${p}_2$ represents their corresponding 3D coordinates. For each pair of 3D-3D correspondences, we compute a scale factor $s$ of the translation vector:
\begin{equation}
    s=\frac{\|{p}_1-{R}{p}_2\|}{\|{\bar{t}}\|},
\label{recovery}
\end{equation}
where $\|{p}_1-{R}{p}_1\|$ denotes the distance difference between 3D points in two images, and $\|\bar{t}\|$ is the length of the translation vector. 
After obtaining the scales between different point pairs, we use the RANSAC algorithm~\cite{fischler1981random} to select the one with the highest agreement as the final scale factor. 
Finally, we transform the scale-free translation vector $\bar{t}$ to the scaled translation vector $t = s \times \bar{t}$.
Combining $t$ with the rotation matrix $R$ to form the final relative pose of the query image with respect to the reference image. 
Combining the translation vector $t$ with the rotation matrix $R$ yields the final relative pose of the query image with respect to the reference image.
% Our algorithm is detailed in Algorithm \ref{code}.

\begin{algorithm}[h]
\SetAlgoLined
\textbf{Input:} Two input images, $I_1$ and $I_2$.\\ 
\textbf{Output:} Rotation matrix $R$ and translation vector $t$.\\
% Poses ($R$, $t$).\\
\vspace{1mm} \hrule \vspace{1mm}
\tcp{Point Map Reconstruction}
calculate $X_{1,1}$ and $X_{2,1}$ based on  $I_1$ and  $I_2$;\\
\tcp{Instance-enhanced Matching}
$Mask_1, Mask_2 \leftarrow$ $SegGPT(I_1, I_2)$; \\
$Map^{g}=\mathcal{M}(I_1,I_2,X_{1,1},X_{2,1})$;\\
$  Map^{o}=\mathcal{M}(I_1^{o},I_2^{o},X_{1,1}^{o},X_{2,1}^{o})$ \\
calculate $Map$ based on Eq.~\ref{eqmap};\\

\tcp{Pose Estimation} 
compute the rotation matrix $R$ and the scale-free unit vector $\bar{t}$ based on Eq.~\ref{eqE1} and~\ref{eqE2};\\

\tcp{Depth-enhanced Matching}
calculate depth $d_1, d_2$ of $I_1,I_2$ based on  \cite{hu2024metric3d}; \\
% ${(x_1,y_1,z_1),...,(x_n,y_n,z_n)} \leftarrow \mathcal{P}([(x_1,y_1),d_1],...,[(x_n,y_n),d_n])$ (
project matching points into 3D space based on Eq.~\ref{p};\\
% $ t \leftarrow \mathcal{T}{([(x_1,y_1,z_1),(x_2,y_2,z_2),...,(x_n,y_n,z_n)],\bar{t})}$ (
calculate scale $s$ based on Eq.~\ref{recovery}; \\
recovery scaled translation vector $t$;\\
\textbf{Return} $R$, $t$.

\caption{\textsc{Inference}}
\label{code}
\end{algorithm}
% \section*{Acknowledgment}

\section{Experiment}
We evaluate our method using the map-free visual relocalization dataset~\cite{arnold2022map}, following the evaluation protocol outlined by Arnold \textit{et al.}~\cite{arnold2022map}.

\begin{table*}
% \LARGE
\renewcommand\arraystretch{1}
    \centering
    \caption{Quantitative comparisons. Our approach outperforms comparison methods.}
    \label{tab:comparisonWithBaseline}
    % \vspace{2mm}
    \resizebox{\linewidth}{!}{%
    \setlength{\tabcolsep}{3mm}
\scalebox{0.8}{
    \begin{tabular}{l|ccccc}
    \toprule
    \textbf{Method} 
    &\makecell{Average Median \\Translation Error(m)($\downarrow$)}  
    &\makecell{Average Median \\Rotation Error(°)($\downarrow$)}  
    &\makecell{Average Median\\ Reprojection Error(px)($\downarrow$)} 
    &\makecell{Precision\\ @ VCRE $<$ 90px ($\uparrow$)} 
    &\makecell{AUC\\@VCRE $<$ 90px ($\uparrow$)} 
    % &\makecell{AUC \\@ VCRE$<$90px($\downarrow$)} 
    % &\makecell{Average Median \\Pose Error (m, °)($\downarrow$)} 
    % &\makecell{Average Median \\Pose Error (m, °)($\downarrow$)} 
    \\

    \midrule
    SIFT (PnP) & 3.722 &72.291 &207.306 &0.208  & 0.396 \\
    SIFT (Ess.Mat +D.Scale) & 3.331 &74.448 &234.170 &0.187  & 0.395 \\
    LoFTR (PnP) & 2.665 &49.212 &178.491 &0.283   & 0.569 \\
    LoFTR (Ess.Mat + D.Scale) & 2.578 &43.522 &180.556 &0.301  & \underline{0.581} \\
    SuperGlue (Procrustes) & 2.882 &55.455 &210.779 &0.210  & 0.388 \\
    SuperGlue (PnP) & 2.379 &42.269 &173.012 &0.311    & 0.541 \\
    SuperGlue (Ess.Mat +D.Scale) & 2.188 &36.358 &178.542 &0.321  & 0.563 \\

    RPR [ R($\alpha$, $\beta$, $\gamma$) + $\hat{t}$  ] & 2.543 &33.977 &215.479 &0.261    & 0.261 \\
    RPR [ R(q) + t ] & 2.370 &32.390 &169.204  &0.290   & 0.290 \\
    RPR [ R(q) + s $\cdot$ $\hat{t}$ ] & 2.125 &31.772 &172.098 &0.284  & 0.284 \\

    RPR [ R(6D)+t ] & 1.845 &24.172 &163.199 &0.333    & 0.333 \\

    RPR [ R($\alpha$, $\beta$,$ \gamma$) + s $\cdot$ $\hat{t}$($\theta$, $\phi$) ] & 1.764 &25.349 &178.918 &0.336    & 0.336 \\
    
    RPR [ 3D-3D ] & \underline{1.667}  &\underline{22.623} &\underline{158.017} &\underline{0.346}  & 0.346 \\

    \midrule

    Ours &\textbf{ 0.596} &\textbf{9.030} &\textbf{70.439} &\textbf{0.484}  &\textbf{0.591} \\
    \bottomrule
    \end{tabular}
    }}
\end{table*}

\subsection{Main Results}
We analyze current models for insights into better relocalization algorithms. Specifically, we build the comparison method using three components: feature matching methods (SIFT\cite{sift}, LoFTR\cite{sun2021loftr}, and SuperGlue\cite{sarlin2020superglue}), depth estimation (DPT\cite{ranftl2021vision} fine-tuned on KITTI\cite{geiger2012we} and NYUv2\cite{silberman2012indoor}), and pose estimation (5-point solver\cite{nister2004efficient} with MAGSAC++\cite{barath2020magsac++}, PNP\cite{gao2003complete}, and Procrustes\cite{eggert1997estimating}). Additionally, we compare various end-to-end methods, including RPR[3D-3D]. By combining these components and end-to-end methods, we obtain 13 comparison methods. 
% Since we don't have the ground truth of the test set, we implemented these methods on the validation set. The results are presented in Table \ref{tab:comparisonWithBaseline}.

As shown in Table~\ref{tab:comparisonWithBaseline}, the results show that our method achieves a significant reduction in Average Median Pose Error. For example, the RPR[3D-3D] method, which performs best among existing methods, has an Average Median Rotation Error of 22.623°, while our method achieves 9.030°. This improvement is largely due to our advanced feature point matching technique, which retains global matching information while focusing on instance-level matching. This approach reduces significant matching errors and enhances precise local matching.
For Average Median Translation Error, the RPR[3D-3D] method reports an error of 1.667m, whereas our method achieves 0.596m. 
% This improvement stems from our effective feature point matching and precise depth estimation. Our method accurately predicts and processes the original image depth, which is crucial for recovering the translation matrix from the scale-free translation vector, underscoring the importance of precise depth estimation for accurate scale recovery.
For the other three metrics, the Average Median Reprojection Error indicates the overall geometric accuracy of the error. Precision@VCRE $<$ 90px evaluates the success rate under a specific threshold, emphasizing the algorithm's performance within a defined error range. Meanwhile, AUC@VCRE $<$ 90px is a global assessment metric that examines performance across various error thresholds, reflecting the robustness and stability of the algorithm. Our method outperforms all baselines in these metrics, demonstrating the robustness and effectiveness of our approach.

\begin{figure}[h]
  \centering
  \vspace{-5mm}
    \caption{The CDF poses estimation errors across all scenes.}
    \vspace{0.7em} 
    \includegraphics[width=0.7\linewidth]{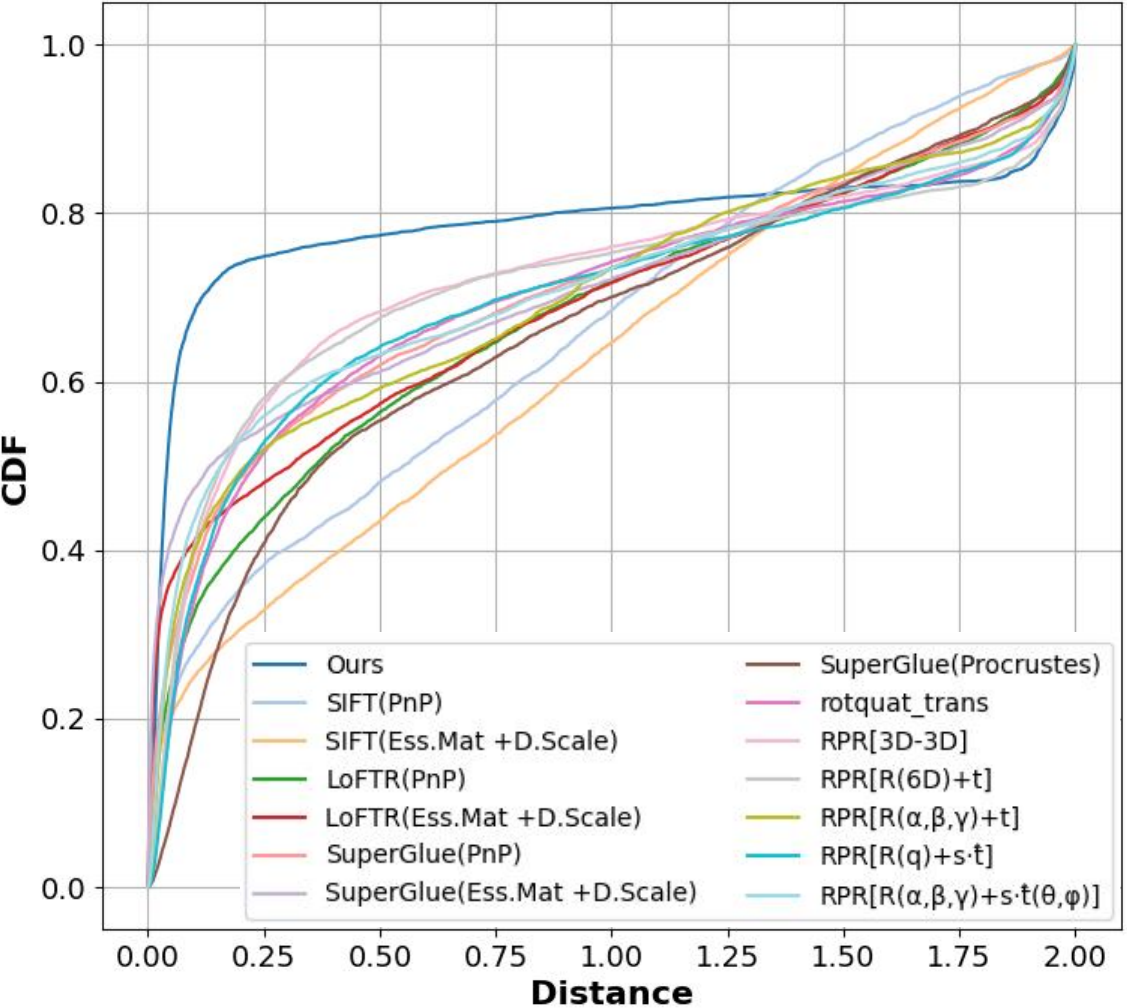}
    \vspace{-5mm} 
    \label{figure:cdf}
\end{figure}

We evaluate our method's effectiveness against the baseline by computing the Cumulative Distribution Function (CDF) for pose estimation errors across all scenes (Figure \ref{figure:cdf}). The CDF illustrates the cumulative probability of errors, with a steeper rise towards 1 at lower error values indicating better performance. Our method's CDF shows a quicker ascent at reduced error levels compared to the baseline, demonstrating its consistently lower pose estimation errors and superior robustness and accuracy across various scenarios.

\subsection{Ablation Study}
In this section, we conduct extensive ablation studies to investigate the significance of instance and depth knowledge within our framework.

\subsubsection{Instance Knowledge} We ablate our instance knowledge and keep all other components unchanged (third line). As shown in Table \ref{tab: Ablation Study}, when instance knowledge is removed, both rotation and translation errors increase. This may be because the absence of accurate feature point matching introduces significant errors when calculating the essential matrix.

\subsubsection{Depth Knowledge}
As directly removing the depth estimation module would cause the model to be unable to output poses, we analyze the importance of depth knowledge by replacing our depth estimation method with DPT~\cite{ranftl2021vision}. Notably, DPT~\cite{ranftl2021vision} has been shown to perform better than other depth estimation models in previous experiments~\cite{arnold2022map}. As shown in the fourth line in Table \ref{tab: Ablation Study}, removing our depth knowledge resulted in a noticeable increase in Median Trans. Error. This is because better depth knowledge enables more accurate scale recovery.

Additionally, we remove both instance and depth knowledge, which is effectively equivalent to DUSt3R~\cite{wang2024dust3r}. As shown in the last line in Table \ref{tab: Ablation Study}, this results in a significant increase in both Median Rot. Error and Median Trans. Error, further emphasizing the importance of instance knowledge and depth knowledge.

\begin{table}[h]
\renewcommand\arraystretch{1.3}
\LARGE
\centering
\vspace{-3mm}
\caption{Ablation study.}
\vspace{-2.5mm}
\resizebox{\linewidth}{!}{
\begin{tabular}{cc|ccc}
% \hline
\toprule
\makecell{Instance} & \makecell{Depth} & \makecell{Median Trans. Error} & \makecell{Median Rot. Error} & \makecell{Median Reproj. Error} \\ 
\midrule
% \hline
\checkmark & \checkmark & 0.596 & 9.030 & 70.439 \\ 
\midrule
& \checkmark & 2.343 & 43.522 & 155.883 \\
\checkmark & & 0.918 & 9.030 & 116.146 \\ 
& & 3.291 & 60.093 & 266.326 \\

\bottomrule
\end{tabular}
\vspace{-8mm}
}

\label{tab: Ablation Study}
\end{table}

\section{Conclusion}We propose a novel map-free relocalization method that estimates the relative pose of a query frame using reference and query images. By leveraging instance segmentation to guide feature point matching within objects, our approach reduces incorrect matches and enhances accuracy. Furthermore, we optimize 3D point coordinates through depth estimation for improved scale recovery. Extensive experiments validate the effectiveness of our collaborative knowledge-enhanced visual relocalization, offering promising advancements in map-free relocalization accuracy.

\clearpage
\bibliographystyle{IEEEtran}
\bibliography{IEEEabrv}
\end{document}